\def\year{2015}
\documentclass[letterpaper]{article}
\usepackage{aaai}
\usepackage{hyperref}
\usepackage{url}
\usepackage{algorithm}
\usepackage{algorithmic}
\usepackage{graphicx}
\usepackage{multirow}
\usepackage{amsmath}
\usepackage{amsfonts}
\usepackage{subfigure}
\usepackage{hyperref}
\usepackage{multirow}
\usepackage{array}
\usepackage{verbatim}
\usepackage{color}
\usepackage{epstopdf}

\usepackage{times}
\usepackage{helvet}
\usepackage{courier}

\usepackage{url}
\usepackage{amsfonts}
\usepackage{booktabs}
\usepackage{subfigure}
\usepackage{array}
\usepackage{verbatim}
\usepackage{caption}

\newcommand{\ignore}[1]{{}}

\def\dup{\tt DUP}
\def\lib{\mathcal{L}}
\def\obs{\mathcal{O}}
\def\a{\mathcal{A}}
\def\uplan{\tilde{p}}
\def\one{\mathbb{I}}
\def\weight{\Gamma}

\title{Discovering Underlying Plans Based on Distributed Representations of Actions}
\author{Xin Tian$^a$, Hankz Hankui Zhuo$^a$ \& Subbarao Kambhampati $^b$\\
  $^a$Sun Yat-Sen University \& $^a$Arizona State University\\
 tianxin1860@gmail.com, zhuohank@mail.sysu.edu.cn, rao@asu.edu}

\begin{document}

\maketitle

\begin{abstract}
Plan recognition aims to discover target plans (i.e., sequences of
actions) behind observed actions, with history plan libraries or
domain models in hand. Previous approaches either discover plans by
maximally ``matching'' observed actions to plan libraries, assuming
target plans are from plan libraries, or infer plans by executing
domain models to best explain the observed actions, assuming complete
domain models are available. In real world applications, however,
target plans are often not from plan libraries and complete domain
models are often not available, since building complete sets of plans
and complete domain models are often difficult or expensive. In this
paper we view plan libraries as corpora and learn vector
representations of actions using the corpora; we then discover target
plans based on the vector representations. Our approach is capable of
discovering underlying plans that are not from plan libraries, without
requiring domain models provided. We empirically demonstrate the
effectiveness of our approach by comparing its performance to
traditional plan recognition approaches in  three planning domains. 
\end{abstract}

\section{Introduction}
As computer-aided cooperative work scenarios become increasingly
popular, human-in-the-loop planning and decision support has become a
critical planning chellenge
(c.f. \cite{cacm-sketch-plan,woogle,ai-mix}).
An important aspect of such a support \cite{aaai-hilp-tutorial} is
recognizing what plans the human in the loop is making, and provide
appropriate suggestions about their next actions.
Although there is a lot of work on plan recognition, much of it has
traditionally depended on the availability of a complete domain model
\cite{geffner-ramirez,hankz-dare}. As has been argued elsewhere
\cite{aaai-hilp-tutorial}, such models are hard to get
in human-in-the-loop planning scenarios. Here, the decision support
systems have to make themselves useful without insisting on complete
action models of the domain. The situation here is akin to that faced
by  search engines and other tools for computer supported
cooperate work, and is thus a significant departure for the ``planning
as pure inference'' mindset of the automated planning community. As
such, the problem 
calls for plan recognition with ``shallow'' models of the domain
(c.f. \cite{rao-model-lite}), that can be easily learned
automatically. 

There has been very little work on learning such shallow models to
support human-in-the-loop planning. Some examples include the work on
Woogle system \cite{woogle} that aimed to provide support to humans in 
web-service composition. That work however relied on very primitive
understanding of the actions (web services in their case) that
consisted merely of learning the input/output types of individual
services. 

In this paper, we focus on learning more informative models that 
that can help recognize the plans under construction by the humans,
and provide active support by suggesting relevant actions. 
To drive this process, we need to learn shallow models of the domain. 
We propose to adapt the recent successes of
word-vector models \cite{word2vec} in language to our problem. Specifically,
we assume that we have access to a corpus of previous plans that the
human user has made. Viewing these plans as made up of action words, we
learn word vector models for these actions. These models provide us a
way to induce the distribution over the identity of each unobserved
action. Given the distributions over individual unobserved actions, we use an
expectation-maximization approach to infer the joint distribution
over all unobserved actions. This distribution then forms the basis for action suggestions. 

We will present the details of our approach, and will also empirically
demonstrate that it does capture a surprising amount of structure in
the observed plan sequences, leading to effective plan recognition. We
further compare its performance to traditional plan recognition
techniques, including one that uses the same plan traces to learn the
STRIPS-style action models, and use the learned model to support plan
recognition. 


\ignore{
We assume that the human is making a plan of at most length $N$. We also
assume that at any given point, the planner is able to observe $N-k$ of
these actions. The $k$ unobserved actions might either be in the suffiix
(i.e., yet to be formed part) of the plan, or in the middle (due to
observational gaps).
Our aim is to suggest, for each of the $k$ unobserved actions, m
possible choices--from which the user can select the action. (Note
that we would like to keep m small, ideally close to 1, so as not to
overwhelm the user)
Accordingly, we will evaluate the effectiveness of the decision
support in terms of whether or not the user's best/intended action is
within the suggested m actions.
}
\section{Problem Definition}
A plan library, denoted by $\lib$, is composed of a set of plans $\{p\}$, where $p$ is a sequence of actions, i.e., $p=\langle a_1,a_2,\ldots,a_n\rangle$ where $a_i$, $1\leq i\leq n$, is an action name (without any parameter) represented by a string. For example, a string \emph{unstack-A-B} is an action meaning that \emph{a robot unstacks block A from block B}. We denote the set of all possible actions by $\bar{\a}$ which is assumed to be known beforehand. For ease of presentation, we assume that there is an empty action, $\O$, indicating an unknown or not observed action, i.e., $\a=\bar{\a}\cup\{\O\}$. An observation of an \emph{unknown} plan $\uplan$ is denoted by $\obs=\langle o_1,o_2,\ldots,o_M\rangle$, where $o_i\in\a$, $1\leq i\leq M$, is either an action in $\bar{\a}$ or an empty action $\O$ indicating the corresponding action is missing or not observed. Note that $\uplan$ is not necessarily in the plan library $\lib$, which makes the plan recognition problem more challenging, since matching the observation to the plan library will not work any more.

We assume that the human is making a plan of at most length $M$. We also
assume that at any given point, the planner is able to observe $M-k$ of
these actions. The $k$ unobserved actions might either be in the suffiix
(i.e., yet to be formed part) of the plan, or in the middle (due to
observational gaps).
Our aim is to suggest, for each of the $k$ unobserved actions, $m$
possible choices--from which the user can select the action. (Note
that we would like to keep $m$ small, ideally close to 1, so as not to
overwhelm the user)
Accordingly, we will evaluate the effectiveness of the decision
support in terms of whether or not the user's best/intended action is
within the suggested $m$ actions.

Specifically, our recognition problem can be represented by a triple $\Re=(\lib,\obs,\a)$. The solution to $\Re$ is to discover the unknown plan $\uplan$ that best explains $\obs$ given $\lib$ and $\a$. An example of our plan recognition problem in the \emph{blocks}\footnote{http://www.cs.toronto.edu/aips2000/} domain is shown below. 

\textbf{Example:} A plan library $\lib$ in the \emph{blocks} domain is assumed to have four plans as shown below:
\begin{center}
\begin{tabular}{|p{0.45\textwidth}|}
\hline
\textbf{plan 1}: \emph{pick-up-B stack-B-A pick-up-D stack-D-C} \\
\textbf{plan 2}: \emph{unstack-B-A put-down-B unstack-D-C put-down-D} \\
\textbf{plan 3}: \emph{pick-up-B stack-B-A pick-up-C stack-C-B pick-up-D stack-D-C} \\
\textbf{plan 4}: \emph{unstack-D-C put-down-D unstack-C-B put-down-C unstack-B-A put-down-B} \\
\hline
\end{tabular}
\end{center}
An observation $\obs$ of action sequence is shown below:
\begin{center}
\begin{tabular}{|p{0.45\textwidth}|}
\hline
\textbf{observation:} \emph{pick-up-B  $\O$ unstack-D-C put-down-D  $\O$ stack-C-B $\O$ $\O$} \\
\hline
\end{tabular}
\end{center}
Given the above input, our {\dup} algorithm outputs plans as follows:
\begin{center}
\begin{tabular}{|p{0.45\textwidth}|}
\hline
\emph{pick-up-B stack-B-A unstack-D-C put-down-D pick-up-C stack-C-B pick-up-D stack-D-C}  \\
\hline
\end{tabular}
\end{center}
\section{Our {\dup} Algorithm}
Our {\dup} approach to the recognition problem $\Re$ functions by two phases. We first learn vector representations of actions using the plan library $\lib$. We then iteratively sample actions for unobserved actions $o_i$ by maximizing the probability of the unknown plan $\uplan$ via the EM framework. We present {\dup} in detail in the following subsections.

\subsection{Learning Vector Representations of Actions}
Since actions are denoted by a name strings, actions can be viewed as words, and a plan can be viewed as a sentence. Furthermore, the plan library $\lib$ can be seen as a corpus, and the set of all possible actions $\a$ is the vocabulary. We thus can learn the vector representations for actions using the Skip-gram model with hierarchical softmax, which has been shown an efficient method for learning high-quality vector representations of words from unstructured corpora \cite{word2vec}. 

The objective of the Skip-gram model is to learn vector representations for predicting the surrounding words in a sentence or document. Given a corpus $\mathcal{C}$, composed of a sequence of training words $\langle w_1,w_2,\ldots,w_T\rangle$, where $T=|\mathcal{C}|$, the Skip-gram model maximizes the average log probability 
\begin{equation}\label{skip-gram}
\frac{1}{T}\sum_{t=1}^T\sum_{-c\leq j\leq c,j\neq0}\log p(w_{t+j}|w_t)
\end{equation} 
where $c$ is the size of the training window or context. 

The basic probability $p(w_{t+j}|w_t)$ is defined by the hierarchical softmax, which uses a binary tree representation of the output layer with the $K$ words as its leaves and for each node, explicitly represents the relative probabilities of its child nodes \cite{word2vec}. For each leaf node, there is an unique path from the root to the node, and this path is used to estimate the probability of the word represented by the leaf node. There are no explicit output vector representations for words. Instead, each inner node has an output vector $v'_{n(w,j)}$, and the probability of a word being the output word is defined by
\begin{eqnarray}\label{prediction}
p(w_{t+j}|w_t)=\prod_{i=1}^{L(w_{t+j})-1}\Big\{\sigma(\one(n(w_{t+j},i+1)= \notag\\ child(n(w_{t+j},i)))\cdot v_{n(w_{t+j},i)}\cdot v_{w_t})\Big\},
\end{eqnarray}
where \[\sigma(x)=1/(1+\exp(-x)).\] $L(w)$ is the length from the root to the word $w$ in the binary tree, e.g., $L(w)=4$ if there are four nodes from the root to $w$. $n(w,i)$ is the $i$th node from the root to $w$, e.g., $n(w,1)=root$ and $n(w,L(w))=w$. $child(n)$ is a fixed child (e.g., left child) of node $n$.  $v_{n}$ is the vector representation of the inner node $n$. $v_{w_t}$ is the input vector representation of word $w_t$. The identity function $\one(x)$ is 1 if $x$ is true; otherwise it is -1.

We can thus build vector representations of actions by maximizing Equation (\ref{skip-gram}) with corpora or plan libraries $\lib$ as input. We will exploit the vector representations to discover the unknown plan $\uplan$ in the next subsection.


\subsection{Maximizing Probability of Unknown Plan $\uplan$}
With the vector representations learnt in the last subsection, a straightforward way to discover the unknown plan $\uplan$ is to explore all possible actions in $\bar{\a}$ such that $\uplan$ has the highest probability, which can be defined similar to Equation (\ref{skip-gram}), i.e.,  
\begin{equation}\label{basic-way}
\mathcal{F}(\uplan)=\sum_{k=1}^M\sum_{-c\leq j\leq c,j\neq0} \log p(w_{k+j}|w_k)
\end{equation}
where $w_k$ denotes the $k$th action of $\uplan$ and $M$ is the length of $\uplan$. As we can see, this approach is exponentially hard with respect to the size of $\bar{\a}$ and number of unobserved actions. We thus design an approximate approach in the Expectation-Maximization framework to estimate an unknown plan $\uplan$ that best explains the observation $\obs$. 

To do this, we introduce new parameters to capture ``weights'' of values for each unobserved action. Specifically speaking, assuming there are $X$ unobserved actions in $\obs$, i.e., the number of $\O$s in $\obs$ is $X$, we denote these unobserved actions by $\bar{a}_1,...,\bar{a}_x,...,\bar{a}_X$, where the indices indicate the order they appear in $\obs$. Note that each $\bar{a}_x$ can be any action in $\bar{\a}$. We associate each possible value of $\bar{a}_x$ with a weight, denoted by $\bar{\weight}_{\bar{a}_x,x}$. $\bar{\weight}$ is a $|\bar{\a}|\times X$ matrix, satisfying \[\sum_{o\in\bar{\a}}\bar{\weight}_{o,x}=1 \wedge \bar{\weight}_{o,x}\geq 0,\] for each $x$. For the ease of specification, we extend $\bar{\weight}$ to a bigger matrix with a size of $|\bar{\a}|\times M$, denoted by $\weight$, such that $\weight_{o,y} = \bar{\weight}_{o,x}$ if $y$ is the index of the $x$th unobserved action in $\obs$, for all $o\in\bar{\a}$; otherwise, $\weight_{o,y} = 1$ and $\weight_{o',y}=0$ for all $o'\in\bar{\a}\wedge o'\neq o$. Our intuition is to estimate the unknown plan $\uplan$ by selecting actions with the highest weights. We thus introduce the weights to Equation (\ref{prediction}), as shown below,
\begin{eqnarray}\label{withpara}
p(w_{k+j}|w_k)=\prod_{i=1}^{L(w_{k+j})-1}\Big\{\sigma(\one(n(w_{k+j},i+1)= \notag\\ child(n(w_{k+j},i)))\cdot a v_{n(w_{k+j},i)}\cdot b v_{w_k})\Big\},
\end{eqnarray}
where $a=\weight_{w_{k+j},k+j}$ and $b=\weight_{w_k,k}$. We can see that the impact of $w_{k+j}$ and $w_k$ is penalized by weights $a$ and $b$ if they are unobserved actions, and stays unchanged, otherwise (since both $a$ and $b$ equal to 1 if they are observed actions). 
\ignore{
Instead of using $\mathcal{F}(\uplan)$, we define a new objective function as shown below, 
\begin{equation}\
\mathcal{G}(\uplan,\weight):=(\prod_{x=1}^X\frac{\weight_{\bar{a}_x,x}}{\sum_{o\in\a,o\neq \bar{a}_x}\weight_{o,x}})\times(\prod_{i=1}^m\prod_{-c\leq j\leq c,j\neq0} p(w_{i+j}|w_i)),
\end{equation}
where $m$ is the length of $\uplan$. In $\mathcal{G}(\uplan,\weight)$, we consider both the likelihood of action $\bar{a}_x$, i.e., $\prod_{x=1}^X\frac{\weight_{\bar{a}_x,x}}{\sum_{o\in\a,o\neq \bar{a}_x}\weight_{o,x}}$, appearing in $\uplan$, and the likelihood of the unknown plan $\uplan$, i.e., $\prod_{i=1}^m\prod_{-c\leq j\leq c,j\neq0} p(w_{i+j}|w_i)$, with restrictions by observed actions in $\uplan$.
}
We redefine the objective function as shown below,
\begin{equation}\label{new-way}
\mathcal{F}(\uplan,\weight)=\sum_{k=1}^M\sum_{-c\leq j\leq c,j\neq0} \log p(w_{k+j}|w_k),
\end{equation}
where $p(w_{k+j}|w_k)$ is defined by Equation (\ref{withpara}). The only parameters needed to be updated are $\weight$, which can be easily done by gradient descent, as shown below,
\begin{equation}\label{update}
\weight_{o,x} = \weight_{o,x} + \delta\frac{\partial\mathcal{F}}{\partial\weight_{o,x}},
\end{equation}
if $x$ is the index of unobserved action in $\obs$; otherwise, $\weight_{o,x}$ stays unchanged, i.e., $\weight_{o,x}=1$. Note that $\delta$ is a learning constant. 

With Equation (\ref{update}), we can design an EM algorithm by repeatedly sampling an unknown plan according to $\weight$ and updating $\weight$ based on Equation (\ref{update}) until reaching convergence (e.g., a constant number of repetitions is reached).

An overview of our {\dup} algorithm is shown in Algorithm \ref{dup}. In Step 2 of Algorithm \ref{dup}, we initialize $\weight_{o,k}=1/M$ for all $o\in\bar{\a}$, if $k$ is an index of unobserved actions in $\obs$; and otherwise, $\weight_{o,k}=1$ and $\weight_{o',k}=0$ for all $o'\in\bar{\a}\wedge o'\neq o$. In Step 4, we view $\weight_{\cdot,k}$ as a probability distribution, and sample an action from $\bar{\a}$ based on $\weight_{\cdot,k}$ if $k$ is an unobserved action index in $\obs$. In Step 5, we only update $\weight_{\cdot,k}$ where $k$ is an unobserved action index. In Step 6, we linearly project all elements of the updated $\weight$ to between 0 and 1, such that we can do sampling directly based on $\weight$ in Step 4. In Step 8, we simply select $\bar{a}_x$ based on \[\bar{a}_x=\arg\max_{o\in\bar{\a}}\weight_{o,x},\] for all unobserved action index $x$.
\begin{algorithm}[!ht]
\caption{Framework of our {\dup} algorithm}\label{dup}
\textbf{Input:} plan library $\lib$, observed actions $\obs$ \\
\textbf{Output:} plan $\uplan$
\begin{algorithmic}[1]
\STATE learn vector representation of actions
\STATE initialize $\weight_{o,k}$ with $1/M$ for all $o\in\bar{\a}$, when $k$ is an unobserved action index
\WHILE{the maximal number of repetitions is not reached} 
	\STATE sample unobserved actions in $\obs$ based on $\weight$
	\STATE update $\weight$ based on Equation (\ref{update})
	\STATE project $\weight$ to [0,1]
\ENDWHILE
\STATE select actions for unobserved actions with the largest weights in $\weight$
\RETURN $\uplan$
\end{algorithmic}
\end{algorithm}

\begin{figure*}[!ht]
\centerline{\includegraphics[width=0.95\textwidth]{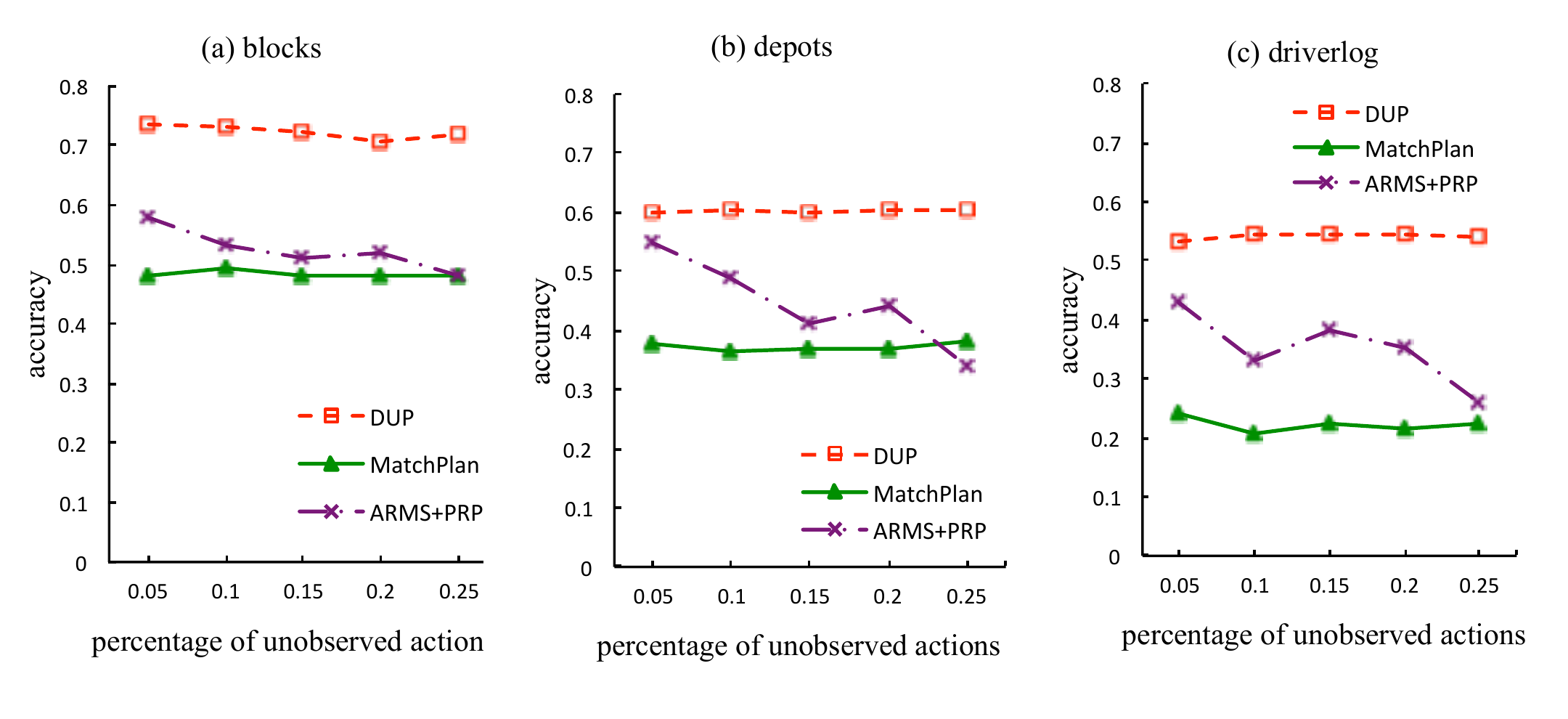}}
\caption{Accuracy with respect to different percentage of unobserved actions}
\label{percent-of-unobs}
\end{figure*}
\section{Experiments}
In this section, we evaluate our {\dup} algorithms in three planning domains from International Planning Competition, i.e., blocks$^1$, depots\footnote{http://www.cs.cmu.edu/afs/cs/project/jair/pub/volume20/long03a-html/JAIRIPC.html}, and driverlog$^2$. To generate training and testing data, we randomly created 5000 planning problems for each domain, and solved these planning problems with a planning solver, such as FF\footnote{https://fai.cs.uni-saarland.de/hoffmann/ff.html}, to produce 5000 plans. We then randomly divided the plans into ten folds, with 500 plans in each fold. We ran our {\dup} algorithm ten times to calculate an average of accuracies, each time with one fold for testing and the rest for training. In the testing data, we randomly removed actions from each testing plan (i.e., $\obs$) with a specific percentage $\xi$ of the plan length.  Features of datasets are shown in Table \ref{dataset}, where the second column is the number of plans generated, the third column is the total number of words (or actions) of all plans, and the last column is the size of vocabulary used in all plans. 
\begin{table}[!th]
\centering
\caption{Features of datasets}\label{dataset}
\begin{tabular}{|l|l|l|l|}
\hline
domain   & \#plan & \#word & \#vocabulary \\\hline
blocks     & 5000 & 292250 & 1250\\\hline
depots     & 5000 & 209711 & 2273\\\hline
driverlog     & 5000 & 179621 & 1441\\\hline
\end{tabular}
\end{table}

We define the accuracy of our {\dup} algorithm as follows. For each unobserved action $\bar{a}_x$ {\dup} suggests a set of possible actions $S_x$ which have the highest value of $\weight_{\bar{a}_x,x}$ for all $\bar{a}_x\in\bar{\a}$. If $S_x$ covers the \emph{truth} action $a_{truth}$, i.e., $a_{truth}\in S_x$, we increase the number of correct suggestions $g$ by 1. We thus define the accuracy $acc$ as shown below:
\[acc = \frac{1}{T}\sum_{i=1}^T\frac{\#\langle correct\textrm{-}suggestions\rangle_i}{K_i},\]
where $T$ is the size of testing set, $\#\langle correct\textrm{-}suggestions\rangle_i$ is the number of correct suggestions for the $i$th testing plan, $K_i$ is the number of unobserved actions in the $i$th testing plan. We can see that the accuracy $acc$ may be influenced by $S_x$. We will test different size of $S_x$ in the experiment.

State-of-the-art plan recognition approaches with plan libraries as input aim at finding a plan from plan libraries to best explain the observed actions \cite{DBLP:conf/ijcai/GeibS07}, which we denote by {\tt MatchPlan}. We develop a \emph{MatchPlan} system based on the idea of \cite{DBLP:conf/ijcai/GeibS07} and compare our {\dup} algorithm to {\tt MatchPlan} with respect to different percentage of unobserved actions $\xi$ and different size of suggestion set $S_x$. Another baseline is action-models based plan recognition approach \cite{cof/ijcai/Ramirez09} (denoted by {\tt PRP}, short for Plan Recognition as Planning). Since we do not have action models as input in our {\dup} algorithm, we exploited the action model learning system {\tt ARMS} \cite{journal/aij/Yang07} to learn action models from the plan library and feed the action models to the {\tt PRP} approach. We call this hybrid plan recognition approach {\tt ARMS+PRP}. To learn action models, {\tt ARMS} requires state information of plans as input. We thus added extra information, i.e., initial state and goal of each plan in the plan library, to {\tt ARMS+PRP}. In addition, {\tt PRP} requires as input a set of candidate goals $\mathcal{G}$ for each plan to be recognized in the testing set, which was also generated and fed to {\tt PRP} when testing. In summary, the hybrid plan recognition approach {\tt ARMS+PRP} has more input information, i.e., initial states and goals in plan library and candidate goals $\mathcal{G}$ for each testing example, than our {\dup} approach.

\begin{figure*}[!ht]
\centerline{\includegraphics[width=0.95\textwidth]{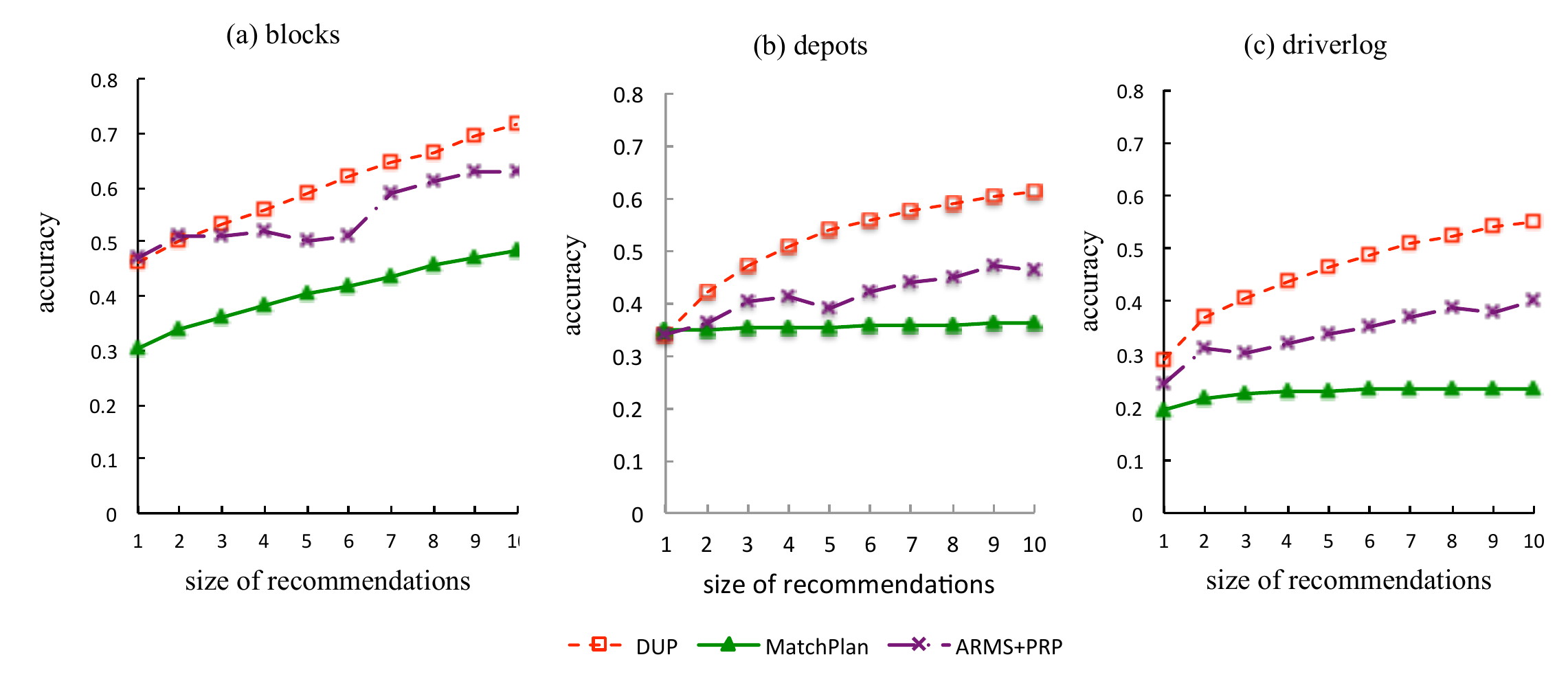}}
\caption{Accuracy with respect to different size of recommendations}
\label{size-of-rec}
\end{figure*}
\subsection{Accuracy w.r.t. Percentage of Unobserved Actions}
We first evaluate our {\dup} algorithm with respect to different percentage of unobserved actions $\xi$ in $\obs$. We set the window of training context $c$ in Equation (\ref{skip-gram}) to be three and the size of recommendations to be ten. We compare our {\dup} algorithm to both {\tt MatchPlan} and {\tt ARMS+PRP}. To make fair comparison (to {\tt MatchPlan}), we set the matching window {\tt MatchPlan} to be three as well when searching plans from plan libraries $\lib$. In other words, to estimate an unobserved action $\bar{a}_x$ in $\obs$, {\tt MatchPlan} matches previous three actions and subsequent three actions of $\bar{a}_x$ to plans in $\lib$, and recommends ten actions with maximal number of matched actions, considering unobserved actions ($\O$ in the context of $\bar{a}_x$) and actions in $\lib$ as a successful matching. For {\tt ARMS+PRP}, we generated 20 candidate goals for each testing example including the ground-truth goal which corresponds to the ground-truth plan to be recognized. The results are shown in Figure \ref{percent-of-unobs}. 

From Figure \ref{percent-of-unobs}, we can see that in all three domains, the accuracy of our {\dup} algorithm is generally higher than {\tt MatchPlan} and {\tt ARMS+PRP}, which verifies that our {\dup} algorithm can indeed capture relations among actions better than previous matching approaches. The rationale is that we explore global plan information from the plan library to learn a ``shallow'' model (distributed representations of actions) and use this model with global information to best explain the observed actions. In contrast, {\tt MatchPlan} just utilizes local plan information when matching the observed actions to the plan library which results in lower accuracies. Although {\tt ARMS+PRP} tries to leverage global plan information from the plan library to learn action models and uses the models to recognize observed actions, it enforces itself to extract ``exact'' models represented by planning models which are often with noise. When feeding those noisy models to {\tt PRP}, since {\tt PRP} that uses planning techniques to recognize plans is very sensitive to noise of planning models, the recognition accuracy is lower than {\dup}, even though {\tt ARMS+PRP} has more input information (i.e., initial states and candidate goals) than our {\dup} algorithm.

Looking at the changes of accuracies with respect to the percentage of
unobserved actions, we can see that our {\dup} algorithm performs
fairly well even when the percentage of unobserved action reaches
25\%. In contrast,  {\tt ARMS+PRP} is sensitive to the percentage of unobserved
actions, i.e., the accuracy goes down when more actions are
unobserved. This is because the noise of planning models induces more
uncertain information, which harms the recognition accuracy, when the
percentage of unobserved actions becomes larger.  Comparing accuracies
of different domains, we can see that our {\dup} algorithm functions
better in the \emph{blocks} domain than the other two domains. This is
because the ratio of \#word over \#vocabulary in the \emph{blocks}
domain is much larger than the other two domains, as shown in Table
\ref{dataset}. We would conjecture that increasing the ratio could
improve the accuracy of {\dup}.

\subsection{Accuracy w.r.t. Size of Recommendation Set}
We next evaluate the performance of our {\dup} algorithm with respect to the size of recommendation set $S_x$. Likewise, we set the context window $c$ used in Equation (\ref{skip-gram}) to be three, which was also set when matching the observed actions $\obs$ to plan libraries $\lib$ in the {\tt MatchPlan} approach. For {\tt ARMS+PRP}, the number of candidate goals for each testing example is set to 20. {\tt ARMS+PRP} aims to recognize plans that are optimal with respect to the cost of actions. We relax {\tt ARMS+PRP} to output $|S_x|$ optimal plans, some of which might be suboptimal.  We varied the number of actions recommended by {\dup} (or \emph{MatchPlan}) from 1 to 10. The results are shown in Figure \ref{size-of-rec}.

From Figure \ref{size-of-rec}, we find that accuracies of the three approaches generally become larger when the size of the recommended action set increases in all three domains. This is consistent with our intuition, since the larger the recommended action set is, the higher the possibility for the \emph{truth} action to be in the recommended action set. We can also see that the accuracy of our {\dup} algorithm are generally larger than both {\tt MatchPlan} and {\tt ARMS+PRP} in all three domains, which verifies that our {\dup} algorithm can indeed better capture relations among actions and thus recognize unobserved actions better than the matching approach {\tt MatchPlan} and the planning model learning approach {\tt ARMS+PRP}. The reason is similar to the one given for Figure \ref{percent-of-unobs} in the previous section. That is, the ``shadow'' model learnt by our {\dup} algorithm is better for recognizing plans than both the ``exact'' planning model learnt by {\tt ARMS} for recognizing plans with planning techniques and the local matching approach {\tt MatchPlan}. On the other hand, we can also see the accuracy of {\tt ARMS+PRP} is generally higher than {\tt MatchPlan}. This verifies that the additional information of initial states and candidate goals exploited by {\tt ARMS+PRP} can indeed help improve the accuracy. Furthermore, the advantage of {\dup} becomes even larger when the size of recommended action set increases, which suggests our vector representation based learning approach can better capture action relations when the size of recommended action set is larger. The possibility of actions correctly recognized by {\dup} becomes much larger than the other two approaches when the size of recommendations increases. 

\section{Related work}
\ignore{Activity recognition with sensors has been a major focus in the area of artificial intelligence. There has been an increasing interest in inferring a user’s activities through low-level sensor modeling. Liao et al. \cite{DBLP:journals/ai/LiaoPFK07} applied a dynamic Bayesian network to estimate a person’s locations and transportation modes from logs of GPS data with relatively precise location information. Bui et al. \cite{DBLP:journals/jair/BuiVW02} introduced an abstract hidden Markov model to infer a person’s goal from camera data in an indoor environment, but it is not clear from the article how action sequences are obtained from camera data. Yin et al. \cite{DBLP:conf/aaai/YinSYL05} explicitly relied on training a location-based sensor model to infer locations from signals; the locations are part of the input that can serve as labels in the training data. To reduce the human labeling effort and cope with the changing signal profiles when the environment changes, Yin et al. \cite{DBLP:conf/percom/YinYN05} dealt with the second issue by transferring the labelled knowledge between time periods. Pan et al. \cite{DBLP:conf/aaai/PanYP07, DBLP:conf/aaai/PanKYP07} proposed to perform location estimation through online co-localization, and apply multi-view learning for migrating the labelled data to a new time period.

Deferent from considering location information, Lester et al. \cite{DBLP:conf/ijcai/LesterCKBH05} propose to build user models for different users and recognize user activities based on the models. They treat all the users equally by simply mixing their data in training. However, different users may behave differently given similar sensor observations. For example, a user may visit the coffee shop for meal and the other just enjoys sitting in its outdoor couches to read research paper. These two users probably observe similar WiFi signals, but their activities are quite personalized. This implies that it may not be appropriate to require all the users to share one common, user-independent activity recognizer. Therefore, Zheng et al. \cite{DBLP:conf/ijcai/ZhengY11} proposed to build a personalized activity recognition model by considering the relations among users. \ignore{Recently, Bulling et al. \cite{DBLP:journals/csur/BullingBS14} provided a tutorial on activity recognition. They discussed the key research challenges that human activity recognition shared with general pattern recognition. When activity recognition is performed indoors and in cities using the widely available Wi-Fi signals, there is much noise and uncertainty. }
Using sensor data as input, Hodges and Pollack designed machine learning-based systems for identifying individuals as they perform routine daily activities such as making coffee \cite{DBLP:conf/huc/HodgesP07}. Liao et al. \cite{DBLP:journals/ai/LiaoPFK07} Proposed to infer user transportation modes from readings of radio-frequency identifiers (RFID) and global positioning systems (GPS). \ignore{Freedman et al. \cite{conf/icaps/Freedman14} explore the application of natural language processing (NLP) techniques, i.e., Latent Dirichlet Allocation topic models, to human skeletal data of plan execution traces obtained from a RGB-D sensor.}
 
Many different applications of activity recognition have been studied by researchers. For example, Pollack et al. \cite{DBLP:journals/ras/PollackBCMOPRT03} show that home-based rehabilitation can be provided for people suffering from traumatic brain injuries by automatically monitoring human activities. Chu et al. \cite{DBLP:journals/jaise/ChuSLK12} present a model of interactive activity recognition to determine the user’s state by interpreting sensor data and/or by explicitly querying the user. The system can be used in an assistive system for persons with cognitive disabilities, which can prompt the user to begin, resume, or end tasks. Zheng et al. \cite{DBLP:conf/iaai/ZhengWGT13} proposed to recognize physical activity from Accelerometer Data Using a Multi-Scale Ensemble Method.

Logic-based recognition aims to explore logically consistent explanations of action observations. 
}
Kautz and Allen proposed an approach to recognizing plans based on parsing observed actions as sequences of subactions and essentially model this knowledge as a context-free rule in an ``action grammar'' \cite{cof/aaai/kautz86}. All actions, plans are uniformly referred to as goals, and a recognizer's knowledge is represented by a set of first-order statements called event hierarchy encoded in first-order logic, which defines abstraction, decomposition and functional relationships between types of events. Lesh and Etzioni further presented methods in scaling up activity recognition to scale up his work computationally \cite{DBLP:conf/ijcai/LeshE95}. They automatically constructed plan-library from domain primitives, which was different from \cite{cof/aaai/kautz86} where the plan library was explicitly represented. In these approaches, the problem of combinatorial explosion of plan execution models impedes its application to real-world domains. Kabanza and Filion \cite{DBLP:conf/ijcai/KabanzaFBI13} proposed an anytime plan recognition algorithm to reduce the number of generated plan execution models based on weighted model counting. These approaches are, however, difficult to represent uncertainty. They offer no mechanism for preferring one consistent approach to another and incapable of deciding whether one particular plan is more likely than another, as long as both of them can be consistent enough to explain the actions observed.\ignore{ Bui et al. \cite{cof/ijcai/Bui03,journal/aij/Geib09} presented approaches to probabilistic plan recognition problems. }

Instead of using a library of plans, Ramirez and Geffner \cite{cof/ijcai/Ramirez09} proposed an approach to solving the plan recognition problem using slightly modified planning algorithms, assuming the action models were given as input. Except previous work \cite{cof/aaai/kautz86,cof/ijcai/Bui03,journal/aij/Geib09,cof/ijcai/Ramirez09} on the plan recognition problem presented in the introduction section, Note that action models can be created by experts or learnt by previous systems, such as {\tt ARMS} \cite{journal/aij/Yang07} and {\tt LAMP} \cite{journal/aij/zhuo10}. Saria and Mahadevan presented a hierarchical multi-agent markov processes as a framework for hierarchical probabilistic plan recognition in cooperative multi-agent systems \cite{conf/ICAPS/Saria04}. Singla and Mooney proposed an approach to abductive reasoning using a first-order probabilistic logic to recognize plans \cite{conf/aaai/Singla11}. Amir and Gal addressed a plan recognition approach to recognizing student behaviors using virtual science laboratories \cite{cof/ijcai/Amir11}. Ramirez and Geffner exploited off-the-shelf classical planners to recognize probabilistic plans \cite{cof/aaai/Ramirez10}.

Early work on human-in-the-loop planning scenarios in automated
planning went under the name of ``mixed-initiative planning''
(e.g. \cite{ferguson1996trains}). An important limitation of that work
was that the humans in the loop were helping the automated planner
(with a complete action model) navigate its search space of plans more
efficiently. In contrast, we are interested in planning technology
that  helping humans develop plans, even in the absence of complete
formal models of the planning domain. While some work in web-service
composition (c.f. \cite{woogle}) did focus on this type of planning
support, they were hobbled by being limited to simple input/output
type comparison. In contrast,  we believe that DUP learns and uses a model
that captures more of the structure of the planning domain (while
still not insisting on complete action models).

While DUP focuses on learning models from plan corpora, some recent
work looked at using crowdsourcing to acquire domain models. 
For example,  Lasecki et
al. \cite{DBLP:conf/cscw/LaseckiSKB13} introduce Legion:AR, which
combines the benefits of automatic and human activity labeling for
robust and deployable activity recognition. The system exploits an
active learning approach \cite{DBLP:conf/aaai/Zhao11} in which
automatic activity recognition is augmented with on-demand activity
labels from the crowd when an observed activity cannot be confidently
classified. By engaging a group of people, Legion:AR is able to label
activities as they occur more reliably than a single person can,
especially in complex domains with multiple actors performing
activities quickly. Lasecki et al. \cite{conf/CHI/Lasecki14} built a
crowdsourcing based system called ARchitect, using the crowd to
capture the dependency structure of the actions that make up
activities. Such crowd-sourcing methods  can complement the
plan-corpus based approach
proposed in DUP. 

\section{Conclusion and Discussion}

In this paper we present a novel plan recognition approach {\dup}
based on vector representation of actions. We first learn the vector
representations of actions from plan libraries using the Skip-gram
model which has been demonstrated to be effective. We then discover
unobserved actions with the vector representations by repeatedly
sampling actions and optimizing the probability of potential plans to
be recognized. We also empirically exhibit the effectiveness of our
approach. 

While we focused on a one-shot recognition task in this paper, in
practice, human-in-the-loop planning will consist of multiple
iterations, with DUP recognizing the plan and suggesting action
addition alternatives; the human making a selection and revising the
plan. The aim is to provide a form of flexible plan completion tool,
akin to auto-completers for search engine queries. To do this
efficiently, we need to make the DUP recognition algorithm ``incremental.''

The word-vector based domain model we developed in this paper provides
interesting contrasts to the standard precondition and effect based
action models used in automated planning community. One of our future
aims is to provide a more systematic comparison of the tradeoffs
offered by these models.  Although we have
focused on the ``plan recognition'' aspects of this model until now,
and assumed that ``planning support'' will be limited to suggesting
potential actions to the humans. In future, we will also consider
``critiquing'' the plans being generated by the humans (e.g. detecting
that an action introduced by the human is not consistent with the
model learned by DUP), and ``explaining/justifying'' the suggestions
generated by humans. Here, we cannot expect causal explanations of the
sorts that can be generated with the help of complete action models
(e.g. \cite{petrie}), and will have to develop justifications
analogous to those used in recommendation systems. 

\newpage
\bibliographystyle{aaai}
\bibliography{aaai16,team,rao}

\begin{thebibliography}{}

\bibitem[\protect\citeauthoryear{Amir and Gal}{2011}]{cof/ijcai/Amir11}
Amir, O., and Gal, Y.~K.
\newblock 2011.
\newblock Plan recognition in virtual laboratories.
\newblock In {\em Proceedings of IJCAI},  2392--2397.

\bibitem[\protect\citeauthoryear{Bui}{2003}]{cof/ijcai/Bui03}
Bui, H.~H.
\newblock 2003.
\newblock A general model for online probabilistic plan recognition.
\newblock In {\em Proceedings of IJCAI},  1309--1318.

\bibitem[\protect\citeauthoryear{Cohen \bgroup et al\mbox.\egroup
  }{2015}]{cacm-sketch-plan}
Cohen, P.~R.; Kaiser, E.~C.; Buchanan, M.~C.; Lind, S.; Corrigan, M.~J.; and
  Wesson, R.~M.
\newblock 2015.
\newblock Sketch-thru-plan: a multimodal interface for command and control.
\newblock {\em Commun. {ACM}} 58(4):56--65.

\bibitem[\protect\citeauthoryear{Dong \bgroup et al\mbox.\egroup
  }{2004}]{woogle}
Dong, X.; Halevy, A.~Y.; Madhavan, J.; Nemes, E.; and Zhang, J.
\newblock 2004.
\newblock Simlarity search for web services.
\newblock In {\em (e)Proceedings of the Thirtieth International Conference on
  Very Large Data Bases, Toronto, Canada, August 31 - September 3 2004},
  372--383.

\bibitem[\protect\citeauthoryear{Ferguson, Allen, and
  Miller}{1996}]{ferguson1996trains}
Ferguson, G.; Allen, J.; and Miller, B.
\newblock 1996.
\newblock Trains-95: Towards a mixed-initiative planning assistant.
\newblock In {\em Proceedings of the Third Conference on Artificial
  Intelligence Planning Systems (AIPS-96)},  70--77.
\newblock Edinburgh, Scotland.

\bibitem[\protect\citeauthoryear{Geib and Goldman}{2009}]{journal/aij/Geib09}
Geib, C.~W., and Goldman, R.~P.
\newblock 2009.
\newblock A probabilistic plan recognition algorithm based on plan tree
  grammars.
\newblock {\em Artificial Intelligence} 173(11):1101--1132.

\bibitem[\protect\citeauthoryear{Geib and
  Steedman}{2007}]{DBLP:conf/ijcai/GeibS07}
Geib, C.~W., and Steedman, M.
\newblock 2007.
\newblock On natural language processing and plan recognition.
\newblock In {\em {IJCAI} 2007, Proceedings of the 20th International Joint
  Conference on Artificial Intelligence, Hyderabad, India, January 6-12, 2007},
   1612--1617.

\bibitem[\protect\citeauthoryear{Kabanza \bgroup et al\mbox.\egroup
  }{2013}]{DBLP:conf/ijcai/KabanzaFBI13}
Kabanza, F.; Filion, J.; Benaskeur, A.~R.; and Irandoust, H.
\newblock 2013.
\newblock Controlling the hypothesis space in probabilistic plan recognition.
\newblock In {\em IJCAI}.

\bibitem[\protect\citeauthoryear{Kambhampati and
  Talamadupula}{2015}]{aaai-hilp-tutorial}
Kambhampati, S., and Talamadupula, K.
\newblock 2015.
\newblock Human-in-the-loop planning and decision support.
\newblock rakaposhi.eas.asu.edu/hilp-tutorial.

\bibitem[\protect\citeauthoryear{Kambhampati}{2007}]{rao-model-lite}
Kambhampati, S.
\newblock 2007.
\newblock Model-lite planning for the web age masses: The challenges of
  planning with incomplete and evolving domain models.
\newblock In {\em Proceedings of the Twenty-Second {AAAI} Conference on
  Artificial Intelligence, July 22-26, 2007, Vancouver, British Columbia,
  Canada},  1601--1605.

\bibitem[\protect\citeauthoryear{Kautz and Allen}{1986}]{cof/aaai/kautz86}
Kautz, H.~A., and Allen, J.~F.
\newblock 1986.
\newblock Generalized plan recognition.
\newblock In {\em Proceedings of AAAI},  32--37.

\bibitem[\protect\citeauthoryear{Lasecki \bgroup et al\mbox.\egroup
  }{2013}]{DBLP:conf/cscw/LaseckiSKB13}
Lasecki, W.~S.; Song, Y.~C.; Kautz, H.~A.; and Bigham, J.~P.
\newblock 2013.
\newblock Real-time crowd labeling for deployable activity recognition.
\newblock In {\em CSCW},  1203--1212.

\bibitem[\protect\citeauthoryear{Lasecki \bgroup et al\mbox.\egroup
  }{2014}]{conf/CHI/Lasecki14}
Lasecki, W.~S.; Weingard, L.; Ferguson, G.; and Bigham, J.~P.
\newblock 2014.
\newblock Finding dependencies between actions using the crowd.
\newblock In {\em Proceedings of CHI},  3095--3098.

\bibitem[\protect\citeauthoryear{Lesh and
  Etzioni}{1995}]{DBLP:conf/ijcai/LeshE95}
Lesh, N., and Etzioni, O.
\newblock 1995.
\newblock A sound and fast goal recognizer.
\newblock In {\em IJCAI},  1704--1710.

\bibitem[\protect\citeauthoryear{Manikonda \bgroup et al\mbox.\egroup
  }{2014}]{ai-mix}
Manikonda, L.; Chakraborti, T.; De, S.; Talamadupula, K.; and Kambhampati, S.
\newblock 2014.
\newblock {AI-MIX:} using automated planning to steer human workers towards
  better crowdsourced plans.
\newblock In {\em Proceedings of the Twenty-Eighth {AAAI} Conference on
  Artificial Intelligence, July 27 -31, 2014, Qu{\'{e}}bec City, Qu{\'{e}}bec,
  Canada.},  3004--3009.

\bibitem[\protect\citeauthoryear{Mikolov \bgroup et al\mbox.\egroup
  }{2013}]{word2vec}
Mikolov, T.; Sutskever, I.; Chen, K.; Corrado, G.~S.; and Dean, J.
\newblock 2013.
\newblock Distributed representations of words and phrases and their
  compositionality.
\newblock In {\em NIPS},  3111--3119.

\bibitem[\protect\citeauthoryear{Petrie}{1992}]{petrie}
Petrie, C.~J.
\newblock 1992.
\newblock Constrained decision revision.
\newblock In {\em Proceedings of the 10th National Conference on Artificial
  Intelligence. San Jose, CA, July 12-16, 1992.},  393--400.

\bibitem[\protect\citeauthoryear{Ram{\'{\i}}rez and
  Geffner}{2009a}]{geffner-ramirez}
Ram{\'{\i}}rez, M., and Geffner, H.
\newblock 2009a.
\newblock Plan recognition as planning.
\newblock In {\em {IJCAI} 2009, Proceedings of the 21st International Joint
  Conference on Artificial Intelligence, Pasadena, California, USA, July 11-17,
  2009},  1778--1783.

\bibitem[\protect\citeauthoryear{Ramirez and
  Geffner}{2009b}]{cof/ijcai/Ramirez09}
Ramirez, M., and Geffner, H.
\newblock 2009b.
\newblock Plan recognition as planning.
\newblock In {\em Proceedings of IJCAI},  1778--1783.

\bibitem[\protect\citeauthoryear{Ramirez and
  Geffner}{2010}]{cof/aaai/Ramirez10}
Ramirez, M., and Geffner, H.
\newblock 2010.
\newblock Probabilistic plan recognition using off-the-shelf classical
  planners.
\newblock In {\em Proceedings of AAAI},  1121--1126.

\bibitem[\protect\citeauthoryear{Saria and
  Mahadevan}{2004}]{conf/ICAPS/Saria04}
Saria, S., and Mahadevan, S.
\newblock 2004.
\newblock Probabilistic plan recognitionin multiagent systems.
\newblock In {\em Proceedings of AAAI}.

\bibitem[\protect\citeauthoryear{Singla and Mooney}{2011}]{conf/aaai/Singla11}
Singla, P., and Mooney, R.
\newblock 2011.
\newblock Abductive markov logic for plan recognition.
\newblock In {\em Proceedings of AAAI},  1069--1075.

\bibitem[\protect\citeauthoryear{Yang, Wu, and
  Jiang}{2007}]{journal/aij/Yang07}
Yang, Q.; Wu, K.; and Jiang, Y.
\newblock 2007.
\newblock Learning action models from plan examples using weighted {MAX-SAT}.
\newblock {\em Artificial Intelligence Journal} 171:107--143.

\bibitem[\protect\citeauthoryear{Zhao, Sukthankar, and
  Sukthankar}{2011}]{DBLP:conf/aaai/Zhao11}
Zhao, L.; Sukthankar, G.; and Sukthankar, R.
\newblock 2011.
\newblock Robust active learning using crowdsourced annotations for activity
  recognition.
\newblock In {\em AAAI workshop}.

\bibitem[\protect\citeauthoryear{Zhuo \bgroup et al\mbox.\egroup
  }{2010}]{journal/aij/zhuo10}
Zhuo, H.~H.; Yang, Q.; Hu, D.~H.; and Li, L.
\newblock 2010.
\newblock Learning complex action models with quantifiers and implications.
\newblock {\em Artificial Intelligence} 174(18):1540--1569.

\bibitem[\protect\citeauthoryear{Zhuo, Yang, and
  Kambhampati}{2012}]{hankz-dare}
Zhuo, H.~H.; Yang, Q.; and Kambhampati, S.
\newblock 2012.
\newblock Action-model based multi-agent plan recognition.
\newblock In {\em Advances in Neural Information Processing Systems 25: 26th
  Annual Conference on Neural Information Processing Systems 2012. Proceedings
  of a meeting held December 3-6, 2012, Lake Tahoe, Nevada, United States.},
  377--385.

\end{thebibliography}

\end{document}